\documentclass[conference]{IEEEtran}
\IEEEoverridecommandlockouts

\usepackage{cite}
\usepackage{amsmath,amssymb,amsfonts}
\usepackage{algorithm}
\usepackage{algpseudocode}
\usepackage{graphicx}
\usepackage{array}
\usepackage{verbatim}
\usepackage{booktabs}
\usepackage{textcomp}
\usepackage{xcolor}
\usepackage{multirow} 
\usepackage{subcaption}
\usepackage{caption}
\def\BibTeX{{\rm B\kern-.05em{\sc i\kern-.025em b}\kern-.08em
    T\kern-.1667em\lower.7ex\hbox{E}\kern-.125emX}}
\begin{document}

\title{Uncertainty-Driven Radar-Inertial Fusion for Instantaneous 3D Ego-Velocity Estimation
}
\author{
    \IEEEauthorblockN{
        Prashant Kumar Rai\IEEEauthorrefmark{1},
        Elham Kowsari\IEEEauthorrefmark{1},
        Nataliya Strokina\IEEEauthorrefmark{2},
        Reza Ghabcheloo\IEEEauthorrefmark{1}
    }
    \IEEEauthorblockA{\IEEEauthorrefmark{1}Faculty of Engineering and Natural Sciences, Tampere University, Tampere, Finland}
    \IEEEauthorblockA{\IEEEauthorrefmark{2}Faculty of Information Technology and Communication Sciences, Tampere University, Tampere, Finland}
    Emails: \{prashant.rai, elham.kowsari, nataliya.strokina, reza.ghabcheloo\}@tuni.fi
}
\maketitle

\begin{abstract}
We present a method for estimating ego-velocity in autonomous navigation by integrating high-resolution imaging radar with an inertial measurement unit. The proposed approach addresses the limitations of traditional radar-based ego-motion estimation techniques by employing a neural network to process complex-valued raw radar data and estimate instantaneous linear ego-velocity along with its associated uncertainty. This uncertainty-aware velocity estimate is then integrated with inertial measurement unit data using an Extended Kalman Filter. The filter leverages the network-predicted uncertainty to refine the inertial sensor’s noise and bias parameters, improving the overall robustness and accuracy of the ego-motion estimation. We evaluated the proposed method on the publicly available Coloradar dataset. Our approach achieves significantly lower error compared to the closest publicly available method, and also outperforms both instantaneous and scan matching-based techniques.
\end{abstract}


\section{INTRODUCTION}
Ego-motion estimation is critical for autonomous navigation~\cite{directegom}, using either proprioceptive sensors (e.g., odometers, IMUs) or exteroceptive sensors (e.g., cameras, LiDAR, radar). While proprioceptive sensors offer reliable short-term odometry, they accumulate drift without external correction. In GNSS-denied environments like indoors, tunnels, or urban canyons, autonomous systems must rely on proprioception, often fused with complementary sensing modalities.

Exteroceptive sensors, like cameras and LiDAR, provide rich spatial information that aids in accurate mapping and localization. However, these sensors suffer in adverse weather or low-visibility conditions, such as heavy rain, fog, or dust. In contrast, radar sensors are known for their robustness under such conditions. Recent advancements in high-resolution millimeter-wave (mmWave) radar technology have enhanced its applicability, enabling robust ego-motion estimation even when optical sensors fail~\cite{tirad, gramme}. Radar sensors provide Doppler velocity measurements of surrounding targets, which can be leveraged for ego-motion estimation. Some approaches focus on instantaneous ego-velocity estimation from a single radar scan~\cite{kellner}, avoiding the need for feature detection and tracking, which are commonly required by registration-based methods~\cite{ndtodom,4DEgo}. However, single-scan radar-based methods assume that all detected targets are stationary, requiring outlier rejection methods such as RANSAC. These assumptions may not hold in highly dynamic environments, leading to degraded accuracy~\cite{graphslam}.

Most radar-based ego-motion methods rely on radar point clouds generated through traditional multi-step processing pipelines, which may lead to reduced data richness or resolution. Our approach addresses these limitations by operating directly on raw, complex-valued radar data rather than relying on pre-processed radar point clouds. Specifically, we introduce a complex-valued deep neural network (CVNN)~\cite{cvnn} that estimates translational ego-velocity directly from complex I/Q (in-phase and quadrature) ADC (analog-to-digital converter ) data. The CVNN preserves phase information, which is typically lost in magnitude-based representations, allowing it to better capture motion-related features.

A key aspect of our method is uncertainty quantification. Unlike conventional networks that provide only a point estimate, our CVNN also predicts an associated uncertainty, which models the heteroscedastic~\cite{Kendall2017WhatUD} noise inherent in radar measurements. This uncertainty-aware output is crucial for sensor fusion, as it allows the system to appropriately weigh the reliability of radar-based velocity estimates. To further improve robustness, we integrate the CVNN’s velocity predictions with an Extended Kalman Filter (EKF), fusing radar data with IMU measurements. The EKF plays a crucial role in bias correction and adaptive sensor fusion: it dynamically adjusts its measurement covariance matrix based on the predicted radar uncertainty and the observed IMU noise characteristics. In practice, when the CVNN reports high uncertainty, the EKF down-weights the radar measurements, relying more on IMU predictions. Conversely, when radar confidence is high, it is given greater importance in the fusion process. This adaptive covariance adjustment ensures that each sensor's contribution is dynamically weighted according to its reliability at any given moment.

The fusion of radar-based ego-velocity estimation with EKF enables more accurate high-frequency estimation. This integration is critical, as IMU-based methods alone suffer from drift over time, particularly in the absence of GNSS corrections. Our approach, therefore, not only leverages radar’s all-weather capability but also compensates for the long-term drift of IMU integration, leading to a highly robust and accurate ego-motion estimation system.

\subsection*{Our Contributions}
\begin{itemize} 
\item We introduce a novel complex-valued neural network that directly processes 3D raw radar complex signals to estimate instantaneous linear ego-velocity. This bypasses the traditional multi-step signal processing without losing informative features in the radar scans.
\item Our method predicts both the ego-velocity and its associated uncertainty by modeling the noise-induced variability in radar measurements. This heteroscedastic uncertainty quantification enhances the reliability of velocity estimates, which is crucial for downstream applications such as sensor fusion. 
\item We integrate radar-based velocity estimates with IMU data using an EKF. The EKF leverages the learned uncertainty from the radar network to dynamically adjust measurement noise, mitigating IMU drift and bias accumulation over time while estimating 3D ego-velocity. 
\item We validate our method on the Coloradar dataset, achieving lower error in velocity component estimation in comparison to traditional instantaneous radar ego-velocity and scan-matching based methods. 
\end{itemize}

\section{RELATED WORK}
Recent advancements in high-resolution millimeter-wave radar sensors have increasingly attracted attention due to their robust performance under adverse weather and low-visibility conditions, where traditional optical sensors such as cameras and LiDAR typically fails~\cite{radarauto, radarodom}.

LiDAR-based~\cite{liosam2020shan} and camera-based odometry~\cite{dso} methods have been extensively explored, achieving high accuracy in ego-motion estimation under favorable conditions. However, millimeter-wave radars offer significant advantages over optical sensors, especially in adverse visibility and weather conditions, prompting recent research into radar-based odometry and simultaneous localization and mapping (SLAM). Many radar odometry and SLAM methods have employed 2D spinning radars, utilizing frame-to-frame registration techniques~\cite{cfear, sdndt, utr, tbvslam, slamrad}. Despite their high angular resolution ($<1^{\circ}$), spinning radars provide only two-dimensional data and lack Doppler velocity measurements.

Automotive System-on-Chip (SoC) radars, on the other hand, offer limited fields of view but can provide both 2D and 3D target measurements along with Doppler velocities~\cite{mimo}. Typically, these radars produce sparse radar point clouds derived from processed raw data samples. Radar-based ego-motion estimation using these radars falls into two main categories: classical registration methods inspired by LiDAR-based techniques, such as Normal Distributions Transform (NDT)\cite{3dndt}, and end-to-end neural network approaches\cite{4drvo, ss4d}. Additionally, some methods utilize full 3D heatmaps obtained directly from 3D MIMO radar~\cite{coloradar}, which represent signal intensities and Doppler velocities across range, azimuth, and elevation dimensions. This representation has been explored for 6-DoF ego-motion estimation using architectures combining 3D CNNs and transformers~\cite{4DEgo}.

Another prominent category involves instantaneous ego-velocity estimation from single radar scans using Doppler measurements. Since a single radar sensor measures only radial velocities, it inherently limits estimation to translational components. To achieve full ego-motion estimation, multiple radar sensors~\cite{kellnermulti} or complementary sensors like IMUs are required~\cite{DoerMFI2020, kramrio}. For instance, Kellner et al.\cite{kellnermulti} employed multiple radar sensors to estimate complete motion. Similarly, Dör et al.\cite{DoerMFI2020} fused radar-derived velocity estimates with IMU measurements using an EKF for complete 3D estimation. Another recent approach~\cite{kramrio} integrated millimeter-wave radar measurements and IMU data through batch optimization over sliding windows. Several other recent studies~\cite{huang2024morephysicalenhancedradarinertialodometry, do2024derodeadreckoningbased} have similarly explored radar-inertial odometry using advanced fusion and optimization techniques. Typically, dynamic targets in radar data are filtered using algorithms such as RANSAC, followed by least-squares optimization~\cite{kellnermulti}. However, these filtering processes can discard valuable information and potentially introduce inaccuracies.

Our approach differs from these previous methods by directly utilizing  complex-valued ADC radar data without intermediate processing steps, as introduced by~\cite{ccaego}. Specifically, we propose a novel complex-valued neural network architecture that predicts instantaneous 3D linear ego-velocity while quantifying uncertainty through covariance estimation. We then fuse these radar-based velocity and uncertainty estimates with IMU measurements using an EKF. This combination addresses the limitations posed by noise and biases in IMU data, ensuring accurate and reliable estimation of the complete ego-motion state.   
\begin{figure*}[t]
\centering
\includegraphics[width=0.75\textwidth]{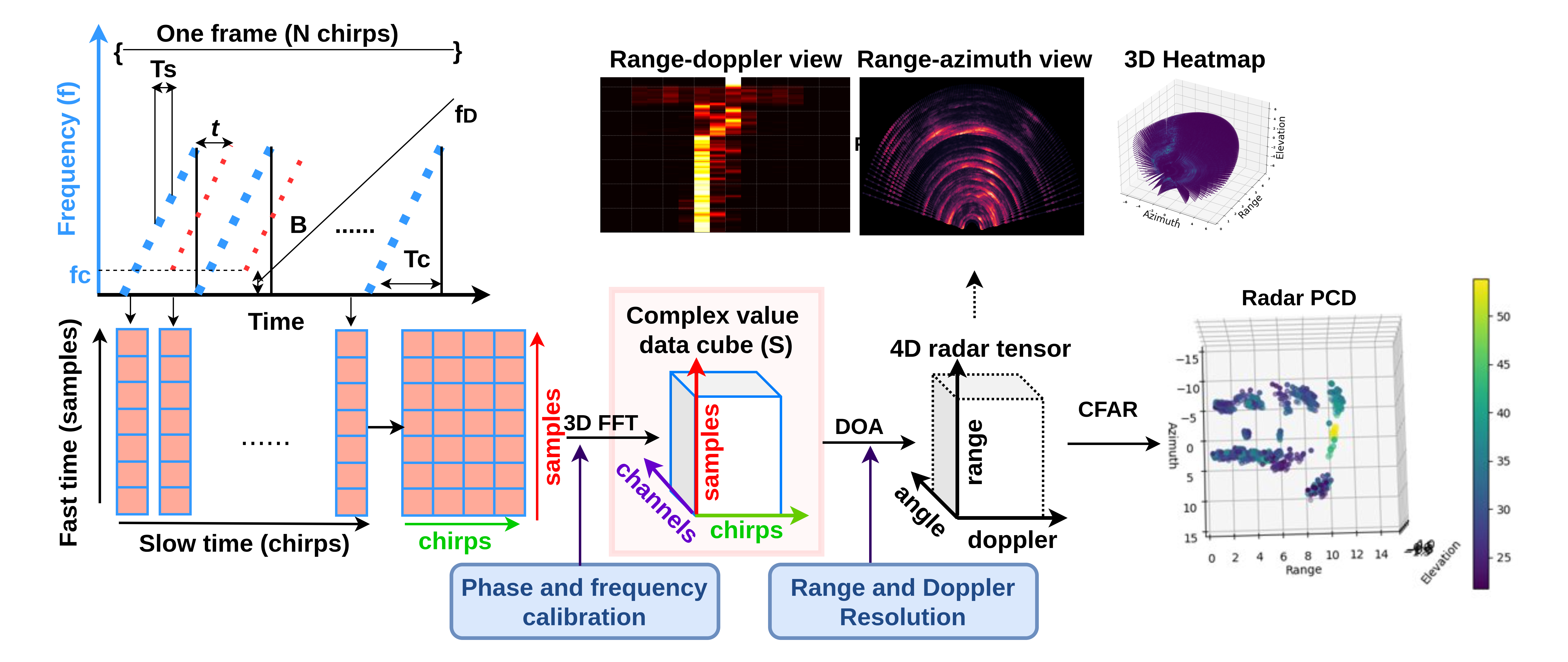}
\caption{TI millimeter-wave (AWR2243) cascade radar signal processing pipeline illustrating the conversion of raw ADC data into a 3D complex-valued data cube via 3D FFT. Our method directly utilizes this minimally processed complex-valued tensor (highlighted block), which encodes range, Doppler, and angular information, as input to the proposed CV-RDCNet.}
\label{fig:datapipe}
\end{figure*}
\section{PROPOSED METHOD}
Our method is loosely coupled fusion method which utilizes the advantages of Doppler radar measurements for linear ego-velocity and IMU for angular ego-velocity resulting full ego-velocity estimation. In this section, we will first introduce our CV-RDCNet (Complex Value - Range Doppler Channel Network) architecture and loss function then the fusion part which includes IMU model and fusion filter.
\begin{figure*}[t]
\centering
\includegraphics[width=0.8\textwidth]{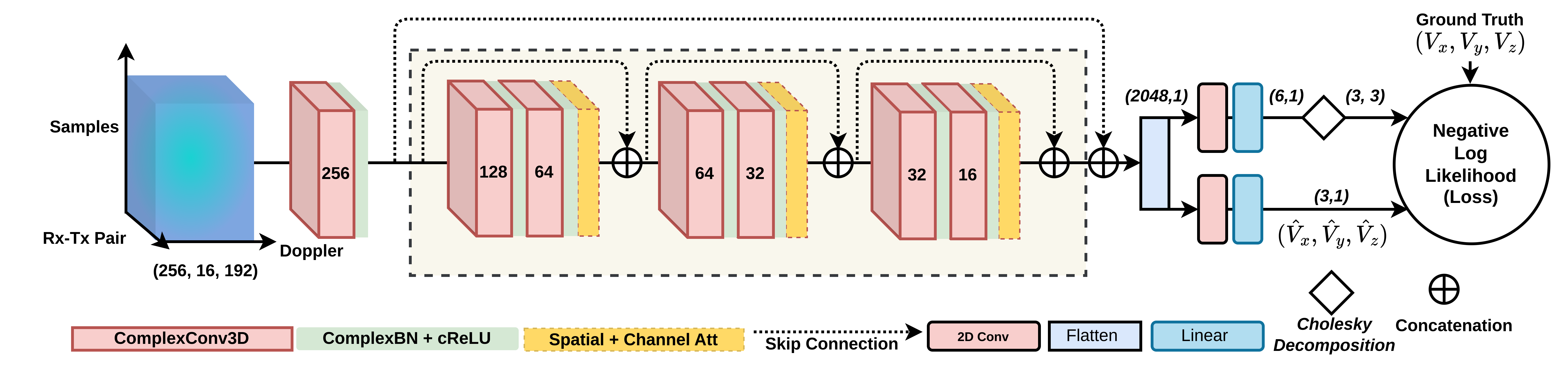}
\caption{Architecture of the proposed CV-RDCNet: a complex-valued convolutional network with attention mechanisms and residual connections for probabilistic ego-velocity estimation from radar data.}
\label{fig:rdcnet}
\end{figure*}

\subsection{Radar CV-RDCNet for linear ego-velocity}
Given a radar input scan \( S \) as shown in Fig.~\ref{fig:datapipe}, our CV-RDCNet as shown in Fig.~\ref{fig:rdcnet} parameterised by weights \(\boldsymbol{\theta}\) predicts the instantaneous linear ego-velocity, represented by a mean vector \(\hat{\mathbf{y}}\in\mathbb{R}^{3}\), and an associated covariance matrix \(\boldsymbol{\Sigma}\in\mathbb{R}^{3\times3}\):

\begin{equation}\label{eq:problem_definition}
    [\hat{\mathbf{y}}, \boldsymbol{\Sigma}] = f_{\boldsymbol{\theta}} (S),
\end{equation}

where \(\hat{\mathbf{y}} = [\hat{V}_x, \hat{V}_y, \hat{V}_z]^\top\) is the predicted mean ego-velocity vector, and \(\boldsymbol{\Sigma}\) quantifies the aleatoric uncertainty associated with these predictions.

Raw radar data (ADC) is represented in complex form, containing phase and frequency information distributed across samples, chirps, and receiver channels. Each dimension respectively represents range bins, Doppler bins, and Tx-Rx channels as shown in Fig~\ref{fig:datapipe}. To effectively manage these complex inputs, complex-valued neural networks have been proposed in the literature~\cite{complexsurvey}. There is also significant work on high-resolution radar data processing~\cite{angleproc} and de-noising techniques~\cite{complex}. However, for applications such as 2D ego-velocity based on complex-valued neural network and channel attention has been explored~\cite{cvca} which shows the significant improvement. Based on the proposed method, we use the CV-RDCNet for 3D ego-velocity estimation using single radar scan and fused the output with IMU. Our network comprises a feature extractor and a state estimation block as illustrated in Fig~\ref{fig:rdcnet}.
The feature extractor in our network includes convolutional residual blocks that feature dual residual complex convolutions enhanced by attention modules. Our end-to-end network processes the complex data cube S (As shown in Fig.~\ref{fig:datapipe}). The extracted features are then flattened and passed through the state estimation block. The feature extractor contains a convolution group with three Complex Residual Blocks (CRBs) followed by a complex convolution layer. Each CRB incorporates a complex-valued convolution, followed by a batch normalization layer, a ReLU activation, and both channel and spatial attention mechanisms.

\subsubsection{Complex Residual Block}
To extract multi-scale features from the radar data, denoted as $S^{R \times D \times A}$, where $R$, $D$, and $A$ represent the range bins ($R=256$), Doppler bins ($D=16$), and angle bins ($A=192$) respectively, the following steps are implemented within our network:

\begin{equation}\label{eq:feature_extraction1}
F1 = ComplexReLU (ComplexBN (ComplexConv (S)))
\end{equation}

In Equation~\ref{eq:feature_extraction1}, the raw data $S$ is first passed through a complex convolution layer (\text{ComplexConv}), followed by complex batch normalization (\text{ComplexBN}), and then activated using the ComplexReLU function. This process initiates the feature extraction by transforming the input data into a more abstract representation.

\begin{equation}\label{eq:feature_extraction2}
F2 = ComplexBN (ComplexConv (F1))
\end{equation}

Equation~\ref{eq:feature_extraction2} further processes the features $F1$ from the previous layer through another complex convolution layer, and the output is normalized using complex batch normalization. This step enhances the stability and efficiency of the network by standardizing the features before they are further processed.

\begin{equation}\label{eq:feature_extraction3}
F3 = SpatialAttention (ChannelAttention (F2))
\end{equation}

In Equation~\ref{eq:feature_extraction3}, an attention mechanism (Spatial + Channel) is applied to $F2$, which allows the network to focus on the most informative features by weighting them based on their significance in the ego-velocity estimation. We use spatial attention on the feature maps (Doppler, Channels) and channel attention on the samples dimension.

Moreover, each complex-valued residual block in the network incorporates a skip connection. This means that the output of each block is concatenated with its input before being passed to the subsequent blocks. This architecture choice helps to mitigate the vanishing gradient problem during training by allowing gradients to flow directly through the network layers, thus enhancing the learning and convergence of the network~\cite{He2015}.

The network is designed to effectively handle the complex-valued input from radar scans, ensuring robust feature extraction for subsequent processing stages.
  
\subsubsection{Linear Ego-motion Prediction with uncertainty}
The features extracted from the radar scan are first down-sampled using strided convolutions to reduce the spatial dimensions of the feature maps for computational efficiency. The down-sampled features passed through a FC (fully connected layer) after flattening to predict the mean of the linear ego-velocity components, \(V_x\), \(V_y\), and \(V_z\). In parallel, a second FC head predicts the parameters required to construct a covariance matrix that represents the uncertainty~\cite{Kendall2017WhatUD, speechenh} in these predictions.

Formally, let \(F_3\) denote the feature maps produced by the preceding layers. The processing is defined as:
\begin{equation}\label{eq:linear_mean}
\hat{\mathbf{y}} = \mathrm{Linear}_{\mu} (flatten (Abs ({StridedConv}_{1\times 1} (F_3))))
\end{equation}
and the covariance matrix is reconstructed as:
\begin{equation}\label{eq:construct_cov}
\boldsymbol{\Sigma} = ConstCov (\mathrm{Linear}_{\Sigma} (flatten (Abs ({SConv}_{1\times 1} (F_3)))))
\end{equation}
where:
\begin{itemize}
    \item $\mathrm{Linear}_{\mu} (\cdot)$ is the FC head that predicts the mean vector $\hat{\mathbf{y}} \in \mathbb{R}^3$,
    \item $\mathrm{Linear}_{\Sigma} (\cdot)$ is the FC head that outputs six parameters from which the covariance matrix $\boldsymbol{\Sigma} \in \mathbb{R}^{3\times 3}$ is reconstructed using a Cholesky decomposition,
    \item $ConstCov (\cdot)$ denotes the transformation from the six predicted parameters to the full covariance matrix.
    \item $SConv (\cdot)$ is the strided convolution layer.
\end{itemize}

Here, the predicted mean $\hat{\mathbf{y}}$ represents the estimated linear ego-velocity components, and the covariance matrix $\boldsymbol{\Sigma}$ provides a measure of aleatoric uncertainty~\cite{KIUREGHIAN2009105, Kendall2017WhatUD} in its predictions.

\subsubsection{Loss Function}\label{subsec:lossfn}
Our network predicts a mean vector $\hat{\mathbf{y}} \in \mathbb{R}^{3}$ and a covariance matrix 
$\boldsymbol{\Sigma} \in \mathbb{R}^{3 \times 3}$ for the three components of linear ego-velocity 
 (i.e., $V_x$, $V_y$, and $V_z$). We assume that the ground truth $\mathbf{y}$ follows a multivariate 
Gaussian distribution:
\begin{equation}
\mathbf{y} \sim \mathcal{N}\left(\hat{\mathbf{y}}, \boldsymbol{\Sigma}\right).
\end{equation}
The corresponding negative log-likelihood (NLL) loss is given by:
\begin{equation}
\mathcal{L}_{\mathrm{NLL}} = \frac{1}{2}\log \lvert \boldsymbol{\Sigma} \rvert + \frac{1}{2} \left(\mathbf{y} - \hat{\mathbf{y}}\right)^\top \boldsymbol{\Sigma}^{-1}\left(\mathbf{y} - \hat{\mathbf{y}}\right)
\end{equation}
A small constant $\epsilon$ is added to the diagonal of $\boldsymbol{\Sigma}$ to ensure numerical stability.

In order to promote meaningful uncertainty estimates and to stop covariance values to getting too smaller, we include additional diagonal regularization.

\textbf{Diagonal Regularization:}  
This term penalizes overly small variances, ensuring that the diagonal elements of $\boldsymbol{\Sigma}$ do not become arbitrarily small~\cite{kumar2019estimating}:
\begin{equation}
    R_{\text{diag}} = \lambda_1 \; \mathbb{E}\left[\frac{1}{\text{diag}(\boldsymbol{\Sigma}) + \epsilon}\right],
\end{equation}
where $\lambda_1$ is a regularization coefficient.

The final loss function is a weighted sum of these components:
\begin{equation}
    \mathcal{L} = \mathcal{L}_{\mathrm{NLL}} + R_{\text{diag}}
\end{equation}

\subsection{ IMU Kinematics and Fusion Filters}
\begin{figure*}[t]  
    \centering
    \begin{subfigure}[b]{0.55\textwidth} 
        \centering
        \includegraphics[width=\linewidth]{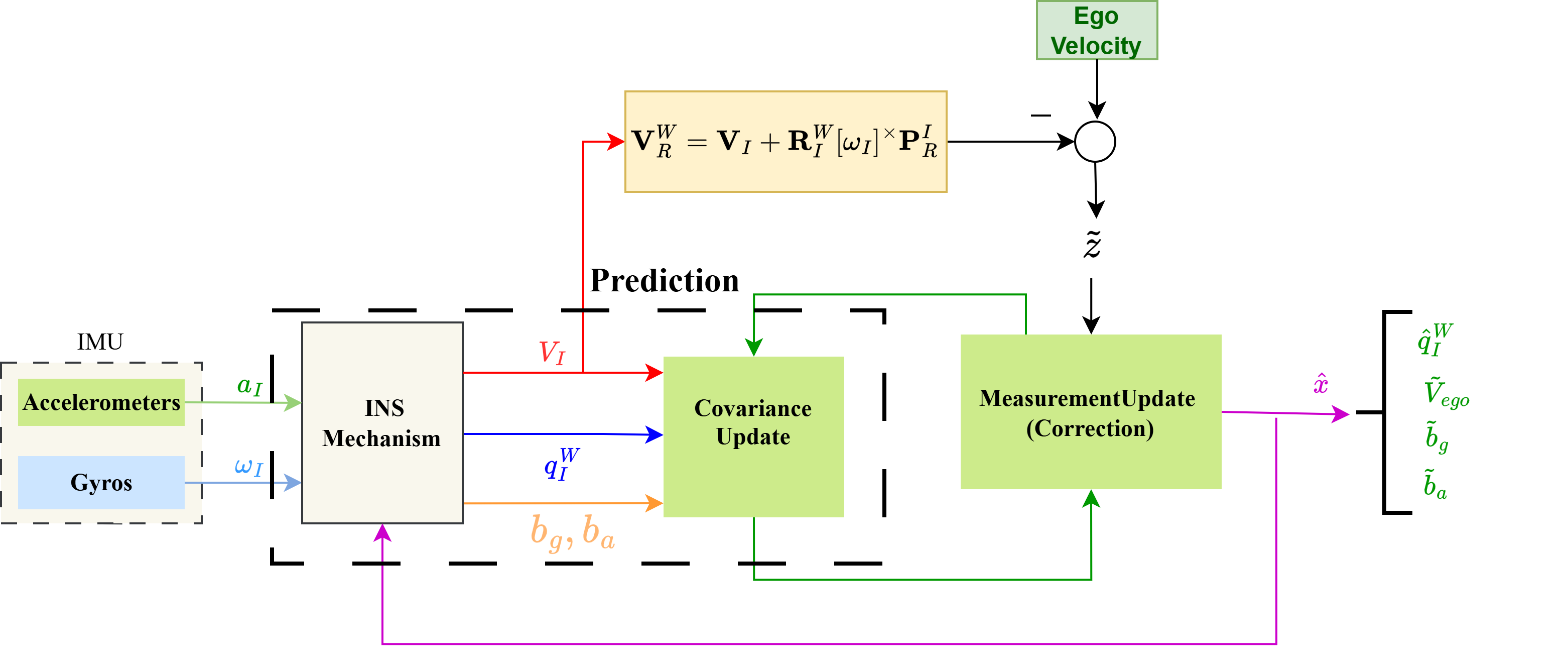} 
        \caption{Radar-Inertial State Estimation using EKF}
        \label{fig:imu}
    \end{subfigure}
    \hspace{0.02\textwidth} 
    \begin{subfigure}[b]{0.30\textwidth} 
        \centering
        \includegraphics[width=\linewidth]{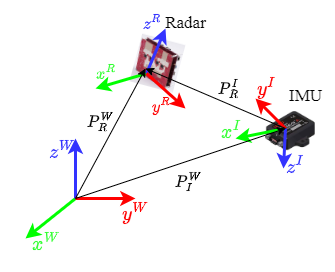} 
        \caption{Coordinate Systems and Relative Positions of IMU and radar}
        \label{fig:sensorbox}
    \end{subfigure}
    
    \caption{Radar-Inertial Sensor Fusion Framework: (a) EKF model for estimating ego-velocity using radar-derived velocity from the proposed CV-RDCNet and IMU measurements to improve accuracy by compensating for the limitations of each sensor. (b) The coordinate system of the radar and IMU, showing their positions and transformations in the world frame for accurate motion estimation.}
    \label{fig:Coordinate}
\end{figure*}

In this section, we present the modeling approach for the IMU and describe how its measurements are fused with the ego-velocity output of the trained CV-RDCNet model (Fig. 3a). Accurate motion estimation is essential for autonomous navigation, particularly when combining data from different sensor modalities. Since inertial sensors such as IMUs are prone to bias and noise, especially over time, estimating these biases is crucial for maintaining the accuracy and reliability of the overall system. To address this, we adopt an Inertial Navigation System (INS) framework that includes bias modelling. The core equations governing the INS mechanism are outlined below:
\begin{equation}\label{eq:IMU_model}
\begin{aligned}
\dot{\mathbf{q}}^W_I &= \tfrac{1}{2} \Omega(\boldsymbol{\omega}_S - \mathbf{b}_g ) \,\mathbf{q}^W_I,\\
\dot{\mathbf{v}}_W &=  {\mathbf{R}^W_I}(\mathbf{a}_S - \mathbf{b}_a) -\mathbf{g}^W,\\
\dot{\mathbf{b}}_g &= w_{b_g,\mathrm{noise}},\\
\dot{\mathbf{b}}_a &= w_{a_g,\mathrm{noise}},
\end{aligned}
\end{equation}
The state vector $\mathbf{x}$ for EKF is defined as:
\begin{equation}
    \mathbf{x} = \begin{bmatrix} q^W_I & \mathbf{V}_I & \mathbf{b}_{\omega} & \mathbf{b}_{a} \end{bmatrix}^T
\end{equation}
where:
\begin{itemize}
    \item $q^W_I$ is the quaternion representing the orientation of the IMU in the world frame, and $\mathbf{R}^W_I$ is the correspondent rotational matrix.
    \item $\mathbf{V}_I$ is the linear velocity of the IMU in the world frame.
    \item $\mathbf{b}_{g}$ is the bias of the gyroscope.
    \item $\mathbf{b}_{a}$ is the accelerometer bias.
 \end{itemize}
    The term $\Omega(\boldsymbol{\omega})$ is the quaternion multiplication matrix, which maps the angular velocity vector to the quaternion space. It is defined as:

\begin{equation}
    \Omega(\boldsymbol{\omega}) =
    \begin{bmatrix}
        0 & -\omega_x & -\omega_y & -\omega_z \\
        \omega_x & 0 & \omega_z & -\omega_y \\
        \omega_y & -\omega_z & 0 & \omega_x \\
        \omega_z & \omega_y & -\omega_x & 0
    \end{bmatrix}
\end{equation}
This matrix is used to compute the derivative of the quaternion, ensuring proper propagation of IMU orientation. We define
\begin{equation}
    \boldsymbol{\delta} = [\,w_{b_g,\mathrm{noise}},\,w_{a_g,\mathrm{noise}}\,]^T
\end{equation}
as zero-mean, uncorrelated Gaussian white noise. In this paper, the world frame is defined as the East-North-Up (ENU) coordinate system and $g^W$ is the gravity vector in this frame.
Since the measurements are the ego-velocities from the radar based on Fig.~\ref{fig:sensorbox} we define the measurement equation as follows:
\begin{equation}\label{Mst_Eq}
\mathbf{V}_R^{W} = \mathbf{V}_I + \mathbf{R}_I^{W } [\mathbf{\omega}_I]^\times \mathbf{P}_R^I
\end{equation}
where $\mathbf{P}_R^I$ represents the transform from the radar frame to the IMU frame.
The Jacobian matrix of the IMU kinematic is shown as follows:
\begin{align}
    F_{k-1} &= \left. \frac{\partial f}{\partial x} \right|_{x = \hat{x}_{k-1}, \delta = \delta_{k-1}} \nonumber \\
            &= \begin{bmatrix}
                \mathbf{0}_{3 \times 3} & \mathbf{0}_{3 \times 3} & -\bar{\mathbf{R}}^W_I & \mathbf{0}_{3 \times 3}\\
                -[\bar{\mathbf{R}}^W_I \mathbf{a}_I]^\times & \mathbf{0}_{3 \times 3} & \mathbf{0}_{3 \times 3} & -\bar{\mathbf{R}}^W_I \\
                \mathbf{0}_{3 \times 3} & \mathbf{0}_{3 \times 3} & \mathbf{0}_{3 \times 3} & \mathbf{0}_{3 \times 3} \\
                \mathbf{0}_{3 \times 3} & \mathbf{0}_{3 \times 3} & \mathbf{0}_{3 \times 3} & \mathbf{0}_{3 \times 3}
            \end{bmatrix}
\label{F_J}
\end{align}

\begin{equation}\
L_{k-1} = \left. \frac{\partial f}{\partial \delta} \right|_{x = \hat{x}_{k-1}} = \begin{bmatrix}
\overline{\mathbf{R}}^W_S & \mathbf{0}_{3 \times 3} & \mathbf{0}_{3 \times 3} & \mathbf{0}_{3 \times 3} \\
\mathbf{0}_{3 \times 3} & \mathbf{I}_{3 \times 3} & \mathbf{0}_{3 \times 3} & \mathbf{0}_{3 \times 3} \\
\mathbf{0}_{3 \times 3} & \mathbf{0}_{3 \times 3} & \mathbf{I}_{3 \times 3} & \mathbf{0}_{3 \times 3} \\
\mathbf{0}_{3 \times 3} & \mathbf{0}_{3 \times 3} & \mathbf{0}_{3 \times 3} & \mathbf{I}_{3 \times 3}
\end{bmatrix}
\end{equation}
and for Jacobin matrix for measurement model (Equation~\ref{Mst_Eq}) the  measurement matrix $H$ is calculated as follows:
\begin{align}
    H_k &= \left. \frac{\partial h}{\partial x} \right|_{x = \hat{x}_k} \nonumber \\
        &= \begin{bmatrix}
            -[\mathbf{R}_I^{W}\mathbf{v}_I]^\times &&& \mathbf{I} &&& -\mathbf{R}_I^{W} [\mathbf{P}_R^I]^\times &&& \mathbf{0}_{3 \times 3}
           \end{bmatrix}
\end{align}
Now, based on the linearized model of our system, we can use the EKF to estimate the linear velocity and biases of the IMU. For the sake of brevity, the estimation algorithm is described as pseudocode in Algorithm \ref{Algo1}.

\begin{algorithm}[!t]
\caption{Estimation Algorithm}\label{Algo1}
\begin{algorithmic}[0] 
\Statex \textbf{Initialize:}
\State $x_0 \sim \mathcal{N}(\hat{x}_0, P_0)$

\Statex \textbf{Prediction:}
\While {no new measurement}
    \State $F_{k-1} = \left. \frac{\partial f}{\partial x} \right|_{x=\hat{x}_{k-1}, \delta=\hat{\delta}_{k-1}}$
    \State $L_{k-1} = \left. \frac{\partial f}{\partial \delta} \right|_{x=\hat{x}_{k-1}}$
    \State $\hat{x}_k = f(\hat{x}_{k-1}, \hat{\delta}_{k-1})$
    \State $P_k = F_{k-1} P_{k-1} F_{k-1}^T + L_{k-1} Q_{k-1} L_{k-1}^T$
\EndWhile

\Statex \textbf{Update:}
\If {new measurement $z_k$ arrives}
    \State $H_k = \left. \frac{\partial h}{\partial x} \right|_{x=\hat{x}_k}$
    \State $\tilde{z}_k = h(\hat{x}_k)$
    \State $K_k = P_k H_k^T(H_k P_k H_k^T + R_k)^{-1}$
    \State $\hat{x}_k = \hat{x}_k + K_k(z_k - \tilde{z}_k)$
    \State $P_k =(I - K_k H_k) P_k$
\EndIf
\end{algorithmic}
\end{algorithm}

\section{EXPERIMENTS}
In order to evaluate the ego motion estimation performance of the proposed method, we first performed the preprocessing to prepare the input radar data and ground truth.
\begin{table}[!t]
    \centering
    \setlength{\tabcolsep}{0.8pt} 
    \renewcommand{\arraystretch}{1.1} 
    \caption{Coloradar data sequences used in our training.}
    \label{tab:datatable}
    \begin{tabular}{l|l|l|l|l|c|c}
        \hline
        \textbf{Environment} & \textbf{Speed} & \textbf{Type} & \textbf{Ground Truth} & \textbf{Platform} & \textbf{\# Seq.} & \textbf{Length (s)} \\
        \hline
        Longboard & fast & outdoor & LiDAR-inertial & e-skateboard & 5 & 170--350 \\
        Edger Army & slow & mine & LiDAR-inertial & walking & 3 & 150--480 \\
        ECR, Hallway & slow & room & LiDAR-inertial & walking & 2 & 100--250 \\
        Outdoor & slow & outdoor & LiDAR-inertial & walking & 6 & 100--200 \\
        \hline
    \end{tabular}
\end{table}
\subsection{Data} 
\subsubsection{Radar Data}
We are using the publicly available Coloradar dataset~\cite{coloradar}, which includes the complex-valued raw ADC data cube from the TI-AWR2243 cascade radar sensor, along with LiDAR, IMU, and ground truth pose trajectories generated in various environments, as referred to in Table~\ref{tab:datatable}. For preprocessing, we performed the 3D-FFT on the raw data to efficiently convert signals from the time domain to the frequency domain, enabling various critical functions such as frequency and phase analysis, range measurement, Doppler processing, and angle estimation, as shown in the processing pipeline in Fig.~\ref{fig:datapipe}. The complex-valued data cube is used as input for training. It is a three-dimensional complex tensor with dimensions representing range, Doppler, and angular domains (Tx-Rx channel dimension). We applied Min-Max normalization to the input complex-valued tensor to scale its values within a consistent range, ensuring stable and efficient training of the neural network. 
\subsubsection{Ground Truth Data}
The dataset includes ground truth poses in the sensor rig frame at 10 frames per second (FPS) in ENU (East-North-Up) frame which we used to calculate the twists using method explained in~\cite{4DEgo} and we used the linear 3d ego-velocity components. For radar-based ego-motion estimation, these ground-truth poses must be in the radar sensor frame. Given that radar operates at a lower frequency (5 FPS), we match ground truth instances with radar timestamps and transform them from the body frame to the sensor frame using the static transform provided in the data set.

\subsection{Training}
We implemented CV-RDCNet using complextorch~\cite{torch}. A data sett of approximately 25,000 instances was used for training. We performed Min-Max normalization on the ground-truth ego-velocities with the preprocessed input data. The network was trained using Adam Optimizer~\cite{adam}, with a learning rate of $10^{-2}$ and a batch size of 128. For training, we used negative log-likelihood loss~\ref{subsec:lossfn}. Training was carried out for 150 epochs with early stop based on validation loss to prevent overfitting. Batch normalization and dropout were used for stable training.

\section{EXPERIMENTAL RESULTS AND ANALYSIS}

To validate the fusion algorithm, we performed an experimental analysis on the Coloradar data set. For evaluation, we selected the data sequences that were not part of the CV-RDCNet training set. For each sequence, we report the mean squared error (MSE) and mean absolute error (MAE).Our fusion algorithm results are systematically compared with those of publicly available baseline approaches.
\begin{itemize}
    \item \textbf{CV-RDCNet + EKF ($ v_m $)}: This method uses the mean velocity ($ v_m $) estimated by CV-RDCNet as the measurement input for the EKF, which combines it with IMU data to estimate ego-motion.
    \item \textbf{CV-RDCNet + EKF ($ v_m, \sigma_m $)}: This method considers both the mean velocity ($ v_m $) and the predicted covariance ($ \sigma_m $) from the neural network. The predicted covariance represents an estimate of heteroscedastic aleatoric uncertainty, allowing the EKF to adjust the measurement confidence dynamically.
\end{itemize}
Uncertainty-aware methods, such as \textbf{CV-RDCNet + EKF ($ v_m, \sigma_m $)}, aim to improve sensor fusion by adjusting the measurement covariance dynamically instead of assuming a fixed noise model. In real-world scenarios, sensor reliability varies due to external factors such as occlusions, multipath reflections, and noise. Estimating uncertainty along with velocity should, in theory, result in better adaptability.

However, despite the expected benefits, the experimental results show that \textbf{CV-RDCNet + EKF ($ v_m $)} slightly outperforms \textbf{CV-RDCNet + EKF ($ v_m, \sigma_m $)}. This can be explained by several factors:
\begin{itemize}
    \item The EKF assumes Gaussian noise, but the uncertainty predicted by the neural network may not follow a Gaussian distribution.
    \item The predicted covariance accounts for epistemic uncertainty, which captures only part of the total uncertainty present in the system.
    \item The estimated uncertainty may not always be accurate, especially in complex and dynamic environments, leading to suboptimal weighting in the EKF.
\end{itemize}

This highlights that while incorporating uncertainty is beneficial for modelling sensor behaviour, improper calibration and assumptions about the underlying uncertainty distribution can lead to suboptimal filtering performance, as discussed in~\cite{pitfallsNLL}.

\subsection{Comparison with Baseline Methods}
We have evaluated the CV-RDCNet along with fusion methods on the test sequences used for evaluation in Table~\ref{tab:results} and compared with the two of existing methods. The results from the model without fusion as shown in Fig.~\ref{fig:predvel}, outperforms the existing models. We indicate that both proposed methods significantly improve ego-motion estimation compared to previous approaches:
\begin{itemize}
    \item RIO~\cite{drio} package is publicly available which we ran on the processed radar scans. It produces higher errors across all the test sequences, showing the limitations of the previous method.
    \item 4DEgo~\cite{4DEgo} which was end-to-end registration method performed on the heatmap data obtained after processing radar scan. It performs better than RIO but is still outperformed by both of our proposed methods.
\end{itemize}

\begin{table*}[t]
    \centering
    \caption{MSE $(\mathrm{m/s})^2$, MAE (m/s), and $\sigma_m$ (for fusion methods) for different test sequences and fusion algorithms (best results in \textbf{Bold}).}
    \label{tab:results}
    \renewcommand{\arraystretch}{1.2}
    \setlength{\tabcolsep}{6pt}
    \begin{tabular}{llccc}
        \toprule
        \textbf{Test Sequence} & \textbf{Algorithm} & \textbf{MSE} & \textbf{MAE} & \textbf{$\sigma_m$} \\
        \midrule
        \multirow{5}{*}{\textbf{Outdoor0}}
        & \textbf{CV-RDCNet + EKF ($v_m$)}           & \textbf{0.0183, 0.0090, 0.0028} & \textbf{0.1023, 0.0749, 0.0382} & \textbf{0.1008, 0.0665, 0.0520} \\
        & \textbf{CV-RDCNet + EKF ($v_m,\sigma_m$)} & \textbf{0.0198, 0.0090, 0.0031} & \textbf{0.1064, 0.0721, 0.0403} & $\sigma_m$ from CV-RDCNet    \\
        & CV-RDCNet ($v_m$)                         & 0.0319, 0.0870, 0.0641         & 0.1786, 0.2949, 0.2532         & --                           \\
        & RIO~\cite{drio}                          & 0.2350, 0.2222, 0.1444         & 0.4848, 0.4714, 0.3800         & --                           \\
        & 4DEgo~\cite{4DEgo}                       & 0.0863, 0.0404, 0.0390         & 0.2076, 0.1421, 0.1396         & --                           \\
        \midrule
        \multirow{5}{*}{\textbf{Longboard2}}
        & \textbf{CV-RDCNet + EKF ($v_m$)}           & \textbf{0.0485, 0.2165, 0.0095} & \textbf{0.1676, 0.3370, 0.0744} & \textbf{0.1824, 0.3842, 0.0948} \\
        & \textbf{CV-RDCNet + EKF ($v_m,\sigma_m$)} & \textbf{0.0495, 0.2255, 0.0103} & \textbf{0.1724, 0.3454, 0.0782} & $\sigma_m$ from CV-RDCNet    \\
        & CV-RDCNet ($v_m$)                         & 0.0927, 0.2621, 0.0893         & 0.2352, 0.3740, 0.2174         & --                           \\
        & RIO~\cite{drio}                          & 1.0185, 10.6338, 0.3763        & 1.0092, 3.2609, 0.4995         & --                           \\
        & 4DEgo~\cite{4DEgo}                       & 0.1281, 0.3108, 0.0692         & 0.2809, 0.4670, 0.2120         & --                           \\
        \midrule
        \multirow{5}{*}{\textbf{Longboard5}}
        & \textbf{CV-RDCNet + EKF ($v_m$)}           & \textbf{0.0753, 0.5141, 0.0133} & \textbf{0.2156, 0.5475, 0.0891} & \textbf{0.2017, 0.6505, 0.1248} \\
        & \textbf{CV-RDCNet + EKF ($v_m,\sigma_m$)} & \textbf{0.0647, 0.5240, 0.0128} & \textbf{0.1978, 0.5577, 0.0891} & $\sigma_m$ from CV-RDCNet    \\
        & CV-RDCNet ($v_m$)                         & 0.2156, 0.6100, 0.1048         & 0.3704, 0.6231, 0.2582         & --                           \\
        & RIO~\cite{drio}                          & 1.2185, 18.6338, 0.2237        & 0.8807, 3.4442, 0.3774         & --                           \\
        & 4DEgo~\cite{4DEgo}                       & 0.1556, 0.4410, 0.0887         & 0.3147, 0.5298, 0.2376         & --                           \\
        \midrule
        \multirow{5}{*}{\textbf{EdgarArmy5}}
        & \textbf{CV-RDCNet + EKF ($v_m$)}           & \textbf{0.0155, 0.0131, 0.0066} & \textbf{0.0983, 0.0798, 0.0585} & \textbf{0.0901, 0.0589, 0.0595} \\
        & \textbf{CV-RDCNet + EKF ($v_m,\sigma_m$)} & \textbf{0.0197, 0.0193, 0.0071} & \textbf{0.1064, 0.1036, 0.0598} & $\sigma_m$ from CV-RDCNet    \\
        & CV-RDCNet ($v_m$)                         & 0.0453, 0.0914, 0.0826         & 0.2128, 0.3023, 0.2874         & --                           \\
        & RIO~\cite{drio}                          & 0.0889, 0.1915, 0.1344         & 0.2378, 0.3491, 0.2925         & --                           \\
        & 4DEgo~\cite{4DEgo}                       & 0.0620, 0.0514, 0.0526         & 0.1986, 0.1808, 0.1829         & --                           \\
        \bottomrule
    \end{tabular}
\end{table*}

\begin{figure}
    \centering
    \includegraphics[width=1.1\linewidth]{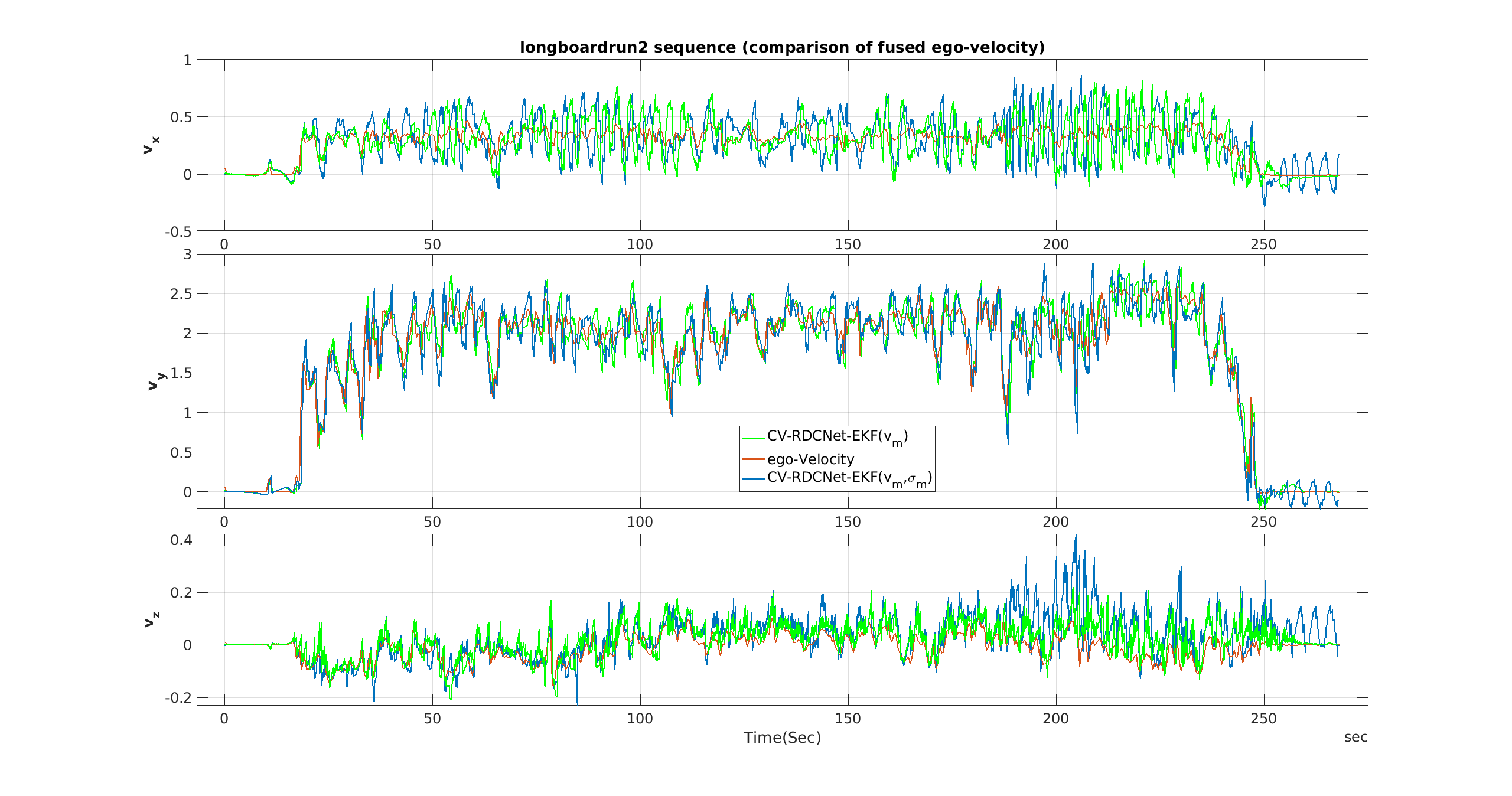}
    \caption{Comparison of estimated ego-velocity across different fusion methods.}
    \label{fig:velocity_comparison}
\end{figure}

\begin{figure}
    \centering
    \includegraphics[width=\linewidth]{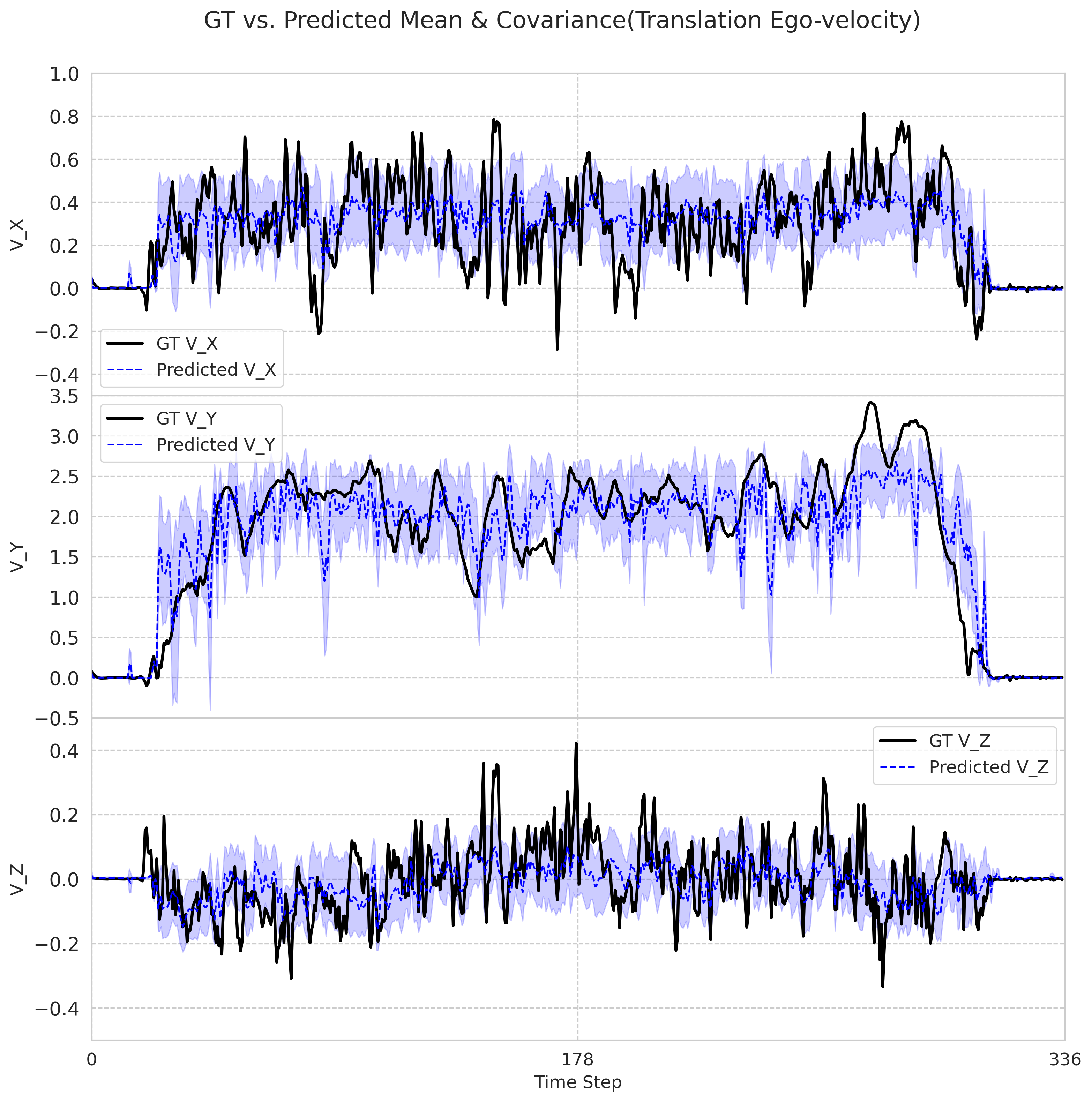}
    \caption{CV-RDCNet predicted mean and sigma vs ground truth for longboardrun2 sequence.}
    \label{fig:predvel}
\end{figure}

\subsection{Analysis of Velocity Estimation}

To further analyse the impact of different fusion strategies on ego-velocity estimation, we visualize the estimated velocities across different methods in Fig.~\ref{fig:velocity_comparison}. The figure compares the performance of \textit{CV-RDCNet + EKF ($v_m$)}, \textit{CV-RDCNet + EKF ($v_m, \sigma_m$)}, and \textit{CV-RDCNet ($v_m$)} over time for the Longboard2 sequence.

\section{CONCLUSION}
This study explored the integration of radar-based ego-velocity estimation with an EKF for robust sensor fusion. The experimental results demonstrated that uncertainty modelling introduces a more flexible approach by allowing dynamic adjustment of measurement confidence. However, the findings indicate that the effectiveness of incorporating predicted uncertainty depends heavily on the assumptions of the filtering framework. The EKF is optimal under Gaussian noise assumptions, while the uncertainty predicted by the neural network may not always conform to this assumption, potentially leading to suboptimal performance.

Furthermore, the results showed that using only the mean velocity estimate ($ v_m $) in the EKF led to slightly better performance than incorporating both the mean velocity and predicted covariance ($ \sigma_m $). This highlights a limitation in the current approach, where the predicted uncertainty does not fully capture all sources of variability in the system. Future work should explore alternative filtering methods, such as the Unscented Kalman Filter (UKF) or Particle Filters (PF), which can better accommodate non-Gaussian uncertainties and improve overall fusion accuracy.

\section*{Acknowledgments}
This project was supported by the Aurora co-innovation project funded by Business Finland and Academy of Finland (project no. 336357, PROFI 6 - TAU Imaging Research Platform, PROFI 7 – SoC and Technology for Intelligent Machines). The authors wish to thank CSC IT Center for Science in Finland for computational resources.


\bibliographystyle{IEEEtran}
\bibliography{reference.bib}

\begin{thebibliography}{10}
\providecommand{\url}[1]{#1}
\csname url@samestyle\endcsname
\providecommand{\newblock}{\relax}
\providecommand{\bibinfo}[2]{#2}
\providecommand{\BIBentrySTDinterwordspacing}{\spaceskip=0pt\relax}
\providecommand{\BIBentryALTinterwordstretchfactor}{4}
\providecommand{\BIBentryALTinterwordspacing}{\spaceskip=\fontdimen2\font plus
\BIBentryALTinterwordstretchfactor\fontdimen3\font minus \fontdimen4\font\relax}
\providecommand{\BIBforeignlanguage}[2]{{%
\expandafter\ifx\csname l@#1\endcsname\relax
\typeout{** WARNING: IEEEtran.bst: No hyphenation pattern has been}%
\typeout{** loaded for the language `#1'. Using the pattern for}%
\typeout{** the default language instead.}%
\else
\language=\csname l@#1\endcsname
\fi
#2}}
\providecommand{\BIBdecl}{\relax}
\BIBdecl

\bibitem{directegom}
C.~Silva and J.~Santos-Victor, ``Direct egomotion estimation,'' in \emph{Proceedings of 13th International Conference on Pattern Recognition}, vol.~1, 1996, pp. 702--706 vol.1.

\bibitem{tirad}
P.~Swami, A.~Jain, P.~Goswami, K.~Chitnis, A.~Dubey, and P.~Chaudhari, ``High performance automotive radar signal processing on ti's tda3x platform,'' in \emph{IEEE Radar Conference (RadarConf)}, 2017.

\bibitem{gramme}
\BIBentryALTinterwordspacing
Y.~Almalioglu, M.~Turan, N.~Trigoni, and A.~Markham, ``Deep learning-based robust positioning for all-weather autonomous driving,'' \emph{Nature Machine Intelligence}, vol.~4, no.~9, pp. 749--760, Sep. 2022. [Online]. Available: \url{https://doi.org/10.1038/s42256-022-00520-5}
\BIBentrySTDinterwordspacing

\bibitem{kellner}
D.~Kellner, M.~Barjenbruch, J.~Klappstein, J.~Dickmann, and K.~Dietmayer, ``Instantaneous ego-motion estimation using doppler radar,'' in \emph{16th International IEEE Conference on Intelligent Transportation Systems (ITSC 2013)}, 2013, pp. 869--874.

\bibitem{ndtodom}
B.~Zhou, Z.~Tang, K.~Qian, F.~Fang, and X.~Ma, ``A lidar odometry for outdoor mobile robots using ndt based scan matching in gps-denied environments,'' in \emph{IEEE International Conference on CYBER Technology in Automation, Control, and Intelligent Systems (CYBER)}, 2017.

\bibitem{4DEgo}
P.~K. Rai, N.~Strokina, and R.~Ghabcheloo, ``4dego: ego-velocity estimation from high-resolution radar data,'' \emph{Frontiers in Signal Processing}, vol.~3, 2023.

\bibitem{graphslam}
W.~Hess, D.~Kohler, H.~Rapp, and D.~Andor, ``Real-time loop closure in 2d lidar slam,'' in \emph{IEEE International Conference on Robotics and Automation (ICRA)}, 2016.

\bibitem{cvnn}
J.~Xu, C.~Wu, S.~Ying, and H.~Li, ``The performance analysis of complex-valued neural network in radio signal recognition,'' \emph{IEEE Access}, vol.~10, pp. 48\,708--48\,718, 2022.

\bibitem{Kendall2017WhatUD}
\BIBentryALTinterwordspacing
A.~Kendall and Y.~Gal, ``What uncertainties do we need in bayesian deep learning for computer vision?'' \emph{ArXiv}, vol. abs/1703.04977, 2017. [Online]. Available: \url{https://api.semanticscholar.org/CorpusID:71134}
\BIBentrySTDinterwordspacing

\bibitem{radarauto}
A.~Srivastav and S.~Mandal, ``Radars for autonomous driving: A review of deep learning methods and challenges,'' \emph{IEEE Access}, vol.~11, pp. 97\,147--97\,168, 2023.

\bibitem{radarodom}
N.~J. Abu-Alrub and N.~A. Rawashdeh, ``Radar odometry for autonomous ground vehicles: A survey of methods and datasets,'' \emph{IEEE Transactions on Intelligent Vehicles}, vol.~9, no.~3, pp. 4275--4291, 2024.

\bibitem{liosam2020shan}
T.~Shan, B.~Englot, D.~Meyers, W.~Wang, C.~Ratti, and R.~Daniela, ``Lio-sam: Tightly-coupled lidar inertial odometry via smoothing and mapping,'' in \emph{IEEE/RSJ International Conference on Intelligent Robots and Systems (IROS)}, 2020.

\bibitem{dso}
J.~Engel, V.~Koltun, and D.~Cremers, ``Direct sparse odometry,'' \emph{IEEE Transactions on Pattern Analysis and Machine Intelligence}, vol.~40, no.~3, pp. 611--625, 2018.

\bibitem{cfear}
D.~Adolfsson, M.~Magnusson, A.~Alhashimi, A.~J. Lilienthal, and H.~Andreasson, ``Lidar-level localization with radar? the cfear approach to accurate, fast, and robust large-scale radar odometry in diverse environments,'' \emph{IEEE Transactions on Robotics}, vol.~39, no.~2, pp. 1476--1495, 2023.

\bibitem{sdndt}
R.~Zhang, Y.~Zhang, D.~Fu, and K.~Liu, ``Scan denoising and normal distribution transform for accurate radar odometry and positioning,'' \emph{IEEE Robotics and Automation Letters}, vol.~8, no.~3, pp. 1199--1206, 2023.

\bibitem{utr}
D.~Barnes and I.~Posner, ``Under the radar: Learning to predict robust keypoints for odometry estimation and metric localisation in radar,'' in \emph{2020 IEEE International Conference on Robotics and Automation (ICRA)}, 2020, pp. 9484--9490.

\bibitem{tbvslam}
D.~Adolfsson, M.~Karlsson, V.~Kubelka, M.~Magnusson, and H.~Andreasson, ``Tbv radar slam – trust but verify loop candidates,'' \emph{IEEE Robotics and Automation Letters}, vol.~8, no.~6, pp. 3613--3620, 2023.

\bibitem{slamrad}
M.~Holder, S.~Hellwig, and H.~Winner, ``Real-time pose graph slam based on radar,'' in \emph{IEEE Intelligent Vehicles Symposium (IV)}, 2019.

\bibitem{mimo}
F.~Engels, P.~Heidenreich, A.~M. Zoubir, F.~K. Jondral, and M.~Wintermantel, ``Advances in automotive radar: A framework on computationally efficient high-resolution frequency estimation,'' \emph{IEEE Signal Processing Magazine}, 2017.

\bibitem{3dndt}
M.~Magnusson, A.~Lilienthal, and T.~Duckett, ``Scan registration for autonomous mining vehicles using 3d-ndt,'' \emph{Journal of Field Robotics}, 2007.

\bibitem{4drvo}
G.~Zhuoins, S.~Lu, L.~Xiong, H.~Zhouins, L.~Zheng, and M.~Zhou, ``4drvo-net: Deep 4d radar–visual odometry using multi-modal and multi-scale adaptive fusion,'' \emph{IEEE Transactions on Intelligent Vehicles}, pp. 1--15, 2023.

\bibitem{ss4d}
H.~Zhou, S.~Lu, and G.~Zhuo, ``Self-supervised 4-d radar odometry for autonomous vehicles,'' in \emph{2023 IEEE 26th International Conference on Intelligent Transportation Systems (ITSC)}, 2023, pp. 764--769.

\bibitem{coloradar}
A.~Kramer, K.~Harlow, C.~Williams, and C.~Heckman, ``Coloradar: The direct 3d millimeter wave radar dataset,'' \emph{The International Journal of Robotics Research}, 2022.

\bibitem{kellnermulti}
D.~Kellner, M.~Barjenbruch, J.~Klappstein, J.~Dickmann, and K.~Dietmayer, ``Instantaneous ego-motion estimation using multiple doppler radars,'' in \emph{2014 IEEE International Conference on Robotics and Automation (ICRA)}, 2014, pp. 1592--1597.

\bibitem{DoerMFI2020}
C.~Doer and G.~F. Trommer, ``An ekf based approach to radar inertial odometry,'' in \emph{2020 IEEE International Conference on Multisensor Fusion and Integration for Intelligent Systems (MFI)}, 2020, pp. 152--159.

\bibitem{kramrio}
A.~Kramer, C.~Stahoviak, A.~Santamaria, A.-a. Agha-mohammadi, and C.~Heckman, ``Radar-inertial ego-velocity estimation for visually degraded environments,'' in \emph{2020 IEEE International Conference on Robotics and Automation (ICRA)}, 05 2020, pp. 5739--5746.

\bibitem{huang2024morephysicalenhancedradarinertialodometry}
\BIBentryALTinterwordspacing
Q.~Huang, Y.~Liang, Z.~Qiao, S.~Shen, and H.~Yin, ``Less is more: Physical-enhanced radar-inertial odometry,'' 2024. [Online]. Available: \url{https://arxiv.org/abs/2402.02200}
\BIBentrySTDinterwordspacing

\bibitem{do2024derodeadreckoningbased}
\BIBentryALTinterwordspacing
H.~V. Do, Y.~H. Kim, J.~H. Lee, M.~H. Lee, and J.~W. Song, ``Dero: Dead reckoning based on radar odometry with accelerometers aided for robot localization,'' 2024. [Online]. Available: \url{https://arxiv.org/abs/2403.05136}
\BIBentrySTDinterwordspacing

\bibitem{ccaego}
H.~Cho, S.~Choi, Y.-R. Cho, and J.~Kim, ``Complex-valued channel attention and application in ego-velocity estimation with automotive radar,'' \emph{IEEE Access}, vol.~PP, pp. 1--1, 01 2021.

\bibitem{complexsurvey}
C.~Lee, H.~Hasegawa, and S.~Gao, ``Complex-valued neural networks: A comprehensive survey,'' \emph{IEEE/CAA Journal of Automatica Sinica}, vol.~9, no.~8, pp. 1406--1426, 2022.

\bibitem{angleproc}
K.~Kaiser, J.~Daugalas, J.~López-Randulfe, A.~Knoll, R.~Weigel, and F.~Lurz, ``Complex-valued neural networks for millimeter wave fmcw-radar angle estimations,'' in \emph{2022 19th European Radar Conference (EuRAD)}, 2022, pp. 145--148.

\bibitem{complex}
A.~Fuchs, J.~Rock, M.~Toth, P.~Meissner, and F.~Pernkopf, ``Complex-valued convolutional neural networks for enhanced radar signal denoising and interference mitigation,'' in \emph{2021 IEEE Radar Conference (RadarConf21)}, 2021, pp. 1--6.

\bibitem{cvca}
H.-W. Cho, S.~Choi, Y.-R. Cho, and J.~Kim, ``Complex-valued channel attention and application in ego-velocity estimation with automotive radar,'' \emph{IEEE Access}, vol.~9, pp. 17\,717--17\,727, 2021.

\bibitem{He2015}
K.~He, X.~Zhang, S.~Ren, and J.~Sun, ``Deep residual learning for image recognition,'' \emph{arXiv preprint arXiv:1512.03385}, 2015.

\bibitem{speechenh}
K.-L. Chen, D.~D.~E. Wong, K.~Tan, B.~Xu, A.~Kumar, and V.~K. Ithapu, ``Leveraging heteroscedastic uncertainty in learning complex spectral mapping for single-channel speech enhancement,'' in \emph{ICASSP 2023 - 2023 IEEE International Conference on Acoustics, Speech and Signal Processing (ICASSP)}, 2023, pp. 1--5.

\bibitem{KIUREGHIAN2009105}
A.~D. Kiureghian and O.~Ditlevsen, ``Aleatory or epistemic? does it matter?'' \emph{Structural Safety}, vol.~31, no.~2, pp. 105--112, 2009.

\bibitem{kumar2019estimating}
A.~Kumar, E.~Marks, W.~Mou, C.~Feng, and X.~Liu, ``Estimating uncertainty with gaussian log-likelihood loss,'' in \emph{Proceedings of the IEEE/CVF International Conference on Computer Vision Workshops}, 2019, pp. 773--782.

\bibitem{torch}
A.~Paszke, S.~Gross, F.~Massa, A.~Lerer, J.~Bradbury, G.~Chanan, T.~Killeen, Z.~Lin, N.~Gimelshein, L.~Antiga, A.~Desmaison, A.~K\"{o}pf, E.~Yang, Z.~DeVito, M.~Raison, A.~Tejani, S.~Chilamkurthy, B.~Steiner, L.~Fang, J.~Bai, and S.~Chintala, \emph{PyTorch: An Imperative Style, High-Performance Deep Learning Library}.\hskip 1em plus 0.5em minus 0.4em\relax Proceedings of the 33rd International Conference on Neural Information Processing Systems, 2019.

\bibitem{adam}
D.~Kingma and J.~Ba, ``Adam: A method for stochastic optimization,'' \emph{International Conference on Learning Representations}, 12 2014.

\bibitem{pitfallsNLL}
\BIBentryALTinterwordspacing
M.~Seitzer, A.~Tavakoli, D.~Antic, and G.~Martius, ``On the pitfalls of heteroscedastic uncertainty estimation with probabilistic neural networks,'' 2022. [Online]. Available: \url{https://arxiv.org/abs/2203.09168}
\BIBentrySTDinterwordspacing

\bibitem{drio}
H.~Chen, Y.~Liu, and Y.~Cheng, ``Drio: Robust radar-inertial odometry in dynamic environments,'' \emph{IEEE Robotics and Automation Letters}, vol.~8, no.~9, pp. 5918--5925, 2023.

\end{thebibliography}

\end{document}